# SemClinBr – a multi-institutional and multi-specialty semantically annotated corpus for Portuguese clinical NLP tasks

Lucas Emanuel Silva e Oliveira[a], Ana Carolina Peters[a], Adalniza Moura Pucca da Silva[a], Caroline P. Gebeluca[a], Yohan Bonescki Gumiel[a], Lilian Mie Mukai Cintho[a], Deborah Ribeiro Carvalho[a], Sadid A. Hasan[b], Claudia Maria Cabral Moro[a]

[a] Health Technology Program, Pontifical Catholic University of Paraná, Curitiba, PR, Brazil
[b] AI Lab, Philips Research North America, Cambridge, MA, USA

**Abstract**

The high volume of research focusing on extracting patient's information from electronic health records (EHR) has led to an increase in the demand for annotated corpora, which are a very valuable resource for both the development and evaluation of natural language processing (NLP) algorithms. The absence of a multi-purpose clinical corpus outside the scope of the English language, especially in Brazilian Portuguese, is glaring and severely impacts scientific progress in the biomedical NLP field. In this study, we developed a semantically annotated corpus using clinical texts from multiple medical specialties, document types, and institutions. We present the following: (1) a survey listing common aspects and lessons learned from previous research, (2) a fine-grained annotation schema which could be replicated and guide other annotation initiatives, (3) a web-based annotation tool focusing on an annotation suggestion feature, and (4) both intrinsic and extrinsic evaluation of the annotations. The result of this work is the SemClinBr, a corpus that has 1,000 clinical notes, labeled with 65,117 entities and 11,263 relations, and can support a variety of clinical NLP tasks and boost the EHR's secondary use for the Portuguese language.

## Graphical abstract

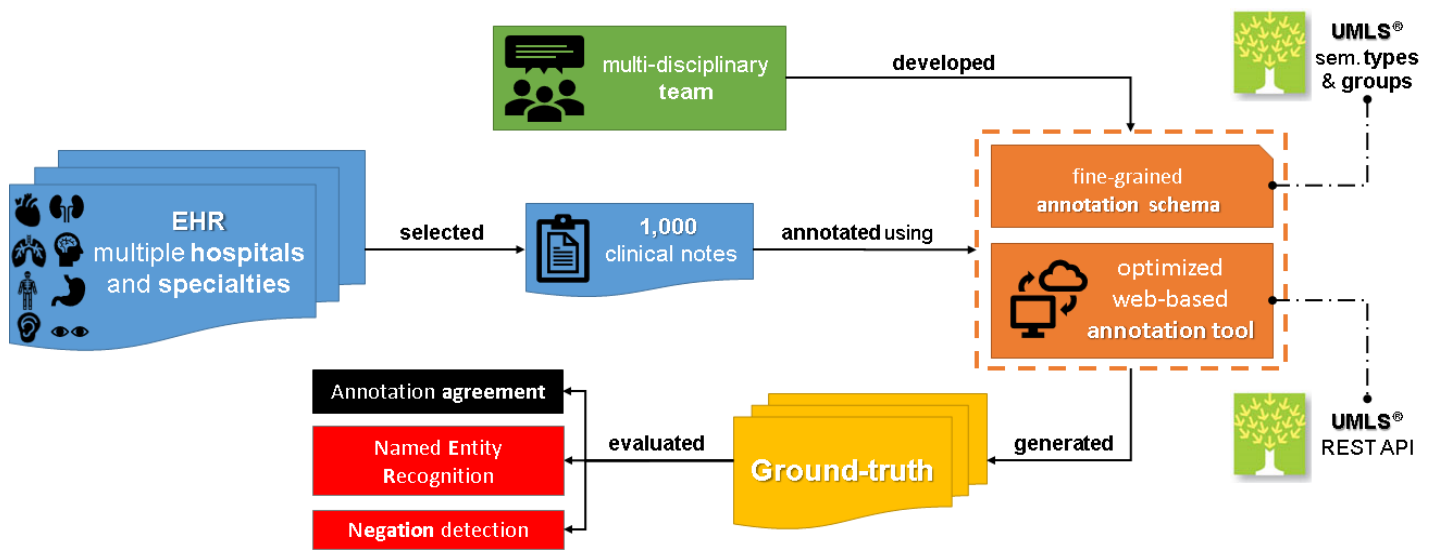

## Highlights

- A survey described the essential aspects and lessons learned regarding biomedical semantic annotation.
- A fine-grained and replicable annotation schema was defined.
- An optimized web-based annotation tool supported the annotation process.

- The SemClinBr corpus consists of 1,000 semantically annotated clinical notes from multiple institutions and medical specialties.
- The intrinsic and extrinsic evaluation of the corpus corroborate its application in different clinical NLP tasks.

# 1. Introduction

In the past two decades, Natural Language Processing (NLP) researchers developed a large amount of work focused on extracting and identifying information among unstructured data (i.e., clinical narratives) stored in the Electronic Health Records (EHR) [1], in what they call the "secondary use of EHR" [2]. Consequently, the scientific community has an increasing demand for corpora with high-quality annotations in order to develop and validate their methods [3]. Semantically annotated corpora can be very useful for both the development and evaluation of NLP and Machine Learning (ML) algorithms aiming to mine information from EHR [4]. When it comes to the clinical domain, this could be a major issue owing to privacy restrictions applied to EHR data: personal health information (PHI) available should not be openly shared for research. Therefore, it is mandatory that we de-identify (anonymize) patient's personal data before we use it, as determined by the Health Insurance Portability and Accountability Act (HIPAA)[1].

Moreover, it is difficult to find a unique corpus that can be potentially applied in several clinical NLP tasks. Such a corpus would be an annotated collection with broad scope and in-depth characteristics, and would, at the same time, present entities with high granularity and comprehensive documents from the point of view of clinical specialties, types, and institutional origin. Working with a language outside the English scope could be another barrier, as most of the annotation work is in English, including the studies developed for shared-tasks and challenges (e.g., [5–12]). Very few initiatives have shared clinical reference corpora in other languages (e.g., [4,13]), and to the best of our knowledge, none of them are in Brazilian Portuguese (pt-br).

Aiming to structure a background to support the biomedical NLP field for pt-br language and address the gaps of broad scope/in-depth clinical corpora outside the English scope, we developed a semantically annotated corpus to assist clinical NLP tasks, both in its evaluation and in its development. We used real clinical texts from multiple institutions, medical specialties, and document types. We defined an annotation schema with fine-granularity entities, described an annotation guidelines document to guide the annotators during the process, and developed a new annotation tool with features that enabled faster and more reliable work. As our main contributions, we highlight (1) the SemClinBR, the first semantically annotated clinical corpus in pt-br; (2) our clinical corpora survey, which lists common steps and lessons learned in corpus development; (3) a replicable annotation schema; and (4) a web-based annotation tool with an annotation assistant incorporated.

# 2. Related work

The use of statistical NLP and ML allowed researchers to automatically retrieve information from biomedical texts and increased the need for gold standard (or ground-truth) corpora to support supervised strategies. Owing to the cost and issues related to annotation projects [14], it is important that the scientific community share its efforts to boost the use of biomedical data and enable researchers to exploit a common evaluation and development environment, which could make a comparison of different methods easier.

The prevalence of biomedical literature corpora over clinical corpora is evident in recent studies [5,14,15]. While the first (normally) deals with open scientific information (e.g., scientific papers, gene data), the second one utilizes EHR personal data, which require, among other things, an anonymization process and ethical committee approval so that the information can be used and releases to the research community. Nevertheless, various clinical semantic annotation initiatives have been developed and shared over the last 10 years; we provide an overview of some of these studies in this section and try to characterize them by listing some of their common features.

The shared-tasks and research challenges are a well-known source of clinical annotated data, as they focus on the development of a specific trending clinical NLP task and provide a common evaluation background for the scientific community. The i2b2 challenges covered important clinical NLP tasks over almost a decade, including clinical data de-identification [16,17], patient smoking status detection [18], obesity and co-morbidities recognition [19], medication extraction [20], concepts/assertions/relation extraction [7], co-reference resolution [21], temporal relation extraction [22], and heart disease risk factors identification [23]. Some of the corpora annotations are described in their own papers [12,24–26] and available to the research community on i2b2 webpage[2]. Another important initiative is the SemEval evaluation series, which focuses on general semantic analysis systems, not only in the biomedical/clinical domain. However, they already shared corpora for specific clinical tasks, such as the "Analysis of Clinical Text" task on SemEval-2014 [10] and SemEval-2015 [11], and the "Clinical TempEval" task on SemEval-2016 [27] and SemEval-2017 [28]. The ShARe/CLEF eHealth labs[3] shared a set of annotated clinical notes for two shared-tasks editions [8,29] and for three different NLP tasks, namely (1) named entity recognition (NER) and normalization of disorders, (2) normalization of acronyms and abbreviations, and (3) patient information retrieval.

---

[1] https://www.hhs.gov/hipaa/for-professionals/privacy/laws-regulations/index.html

To provide a development and evaluation environment for clinical information extraction systems, the CLinical E-Science Framework (CLEF) project built a semantically annotated corpus of clinical texts [5]. The project labeled entities, relations, modifiers, co-reference, and temporal information within the text using a CLEF project tagset. Although its large size (20,234 clinical documents), the corpus focused on patients with neoplasms only. In a recent study, Patel et al., (2018)[30] built a large clinical entity corpus, with 5,160 clinical documents from 40 different medical domains. They annotated a set of 11 semantic groups, which they mapped to the corresponding UMLS semantic types.

The THYME[4] (Temporal Histories of Your Medical Events) corpus [31] is another example of a gold standard produced by annotating clinical notes. The annotation process focused on event and relation annotation, especially with regard to temporal information. Finally, the MiPACQ corpus [32] features syntactic and semantic annotation of clinical narratives (127,606 tokens precisely). The semantic labeling followed the UMLS hierarchy of semantic groups [33] to avoid semantic type ambiguity.

What we have seen so far is a predominance of corpora built for the English language. We did not find any study focusing on clinical semantic annotation for pt-br. However, there is a non-shared corpus in European Portuguese, named the MedAlert Discharge Letters Representation Model (MDLRM) and developed by Ferreira et al. (2010)[34]. They annotated a set of entities (i.e., Condition, Anatomical Site, Evolution, Examination, Finding, Location, Therapeutic, DateTime, and Value) in 90 discharge summaries from a hospital in Portugal aiming to evaluate a NER task.

Furthermore, there are some efforts dedicated to other languages, and we mention some of them below. For Spanish, experts annotated the IxaMed-GS corpus [13] with entities and relations associated with diseases and drugs, using an adaptation of the SNOMED-CT tagset. A notable 3-year-long work is the Medical Entity and Relation LIMSI annOtated Text corpus (MERLOT), which produced a corpus of 500 annotated clinical documents for the French language using an entity annotation scheme[5] partially derived from the UMLS Semantic Groups [4].

A recent study focused on German nephrology reports to build a fine-grained annotated corpus, following a concept type organization similar to the UMLS semantic types/groups. The corpus consists of 118 discharge summaries and 1,607 short evolution notes [35]. Moreover, for the Swedish language, Skeppstedt et al. (2014)[36] annotated a set of highly relevant entities for building a patient overview (Disorder, Finding, Pharmaceutical Drug, and Body Structure) to train a NER algorithm previously applied to English clinical texts. Their corpus has 45,482 tokens in the training set and 25,370 tokens in the evaluation set.

To realize cohesive, reliable, unbiased, and fast annotations, most studies share the following common steps:

- double annotation → to reduce bias and improve reliability
- guidelines/scheme definition → to improve reliability and support annotators
- annotation agreement measures → to ensure reliability
- use of an annotation tool → to ease/speed up the annotation work
- annotation characterization (e.g., semantic types, relations) based on the desired task → for better scope definition

---

[2] https://www.i2b2.org/NLP/DataSets/Main.php
[3] https://sites.google.com/site/shareclefehealth/
[4] https://clear.colorado.edu/TemporalWiki/index.php/Main_Page
[5] https://cabernet.limsi.fr/annotation_guide_for_the_merlot_french_clinical_corpus-Sept2016.pdf

On the other hand, mostly because of annotation costs, issues associated to a high generalization and specificity of the annotation, and difficulties in obtaining clinical data, none of the available corpora shares all the following characteristics together:

- documents from multiple institutions → different writing and care styles
- multiple types of documents (e.g., discharge summaries, nursing notes) → distinct care phases
- documents from various medical specialties (e.g., cardiology, nephrology) → broader clinical view and care perspectives
- multiple medical conditions (e.g., diabetes, cardiovascular disease) →larger dictionary of terms
- high granularity entity annotation (normally they grouped a few entity types) → enables specific labeling
- shared a detailed annotation guideline → allows replication
- a high number of clinical notes → more representativity and ML training conditions
- outside English scope → boost research field in other languages

Although there is a lack of large multi-purpose corpora, it is also necessary to note the need for a heterogeneous clinical corpus for the scientific community. For example, Deleger et al. (2012)[37] argue that most of the clinical data de-identification systems were tested in corpora composed of a unique or small variety of document types (e.g., discharge summaries, nursing notes), when the ideal would be an evaluation using heterogeneous corpora. In addition, when defining the granularity of corpus annotation, one must remember the trade-off between granularity and reliability, as discussed by Crible and Degand (2017)[38], prioritizing each of these aspects according to the objectives of the annotation.

Hovy and Lavid (2010)[39] refer to corpus annotation as "*adding interpretive information into a collection of texts,*" and describe seven main questions in a general annotation project:

1. Representative text selection
2. Concept/Theory instantiation (tagset definition + guidelines first draft)
3. Annotators selection and training (preliminary annotation + guidelines update)
4. Annotation procedure specification (definitive guidelines)
5. Annotation interface design (increase speed and avoid bias)
6. Evaluation measures definition (satisfactory agreement level – in case of low agreement, returns to step 2 – if in good agreement, continue annotation, maintaining intermediate checks, improvements, etc.)
7. Annotation finalization and NLP/ML algorithm deployment

Xia and Yetisgen-Yildiz (2012)[14] listed the challenges and strategies in clinical corpus annotation, and concluded that the medical training of the annotators is not enough on its own to ensure that high-quality annotations will be achieved; as a result, NLP researchers should get involved in the annotation process as early as possible. Moreover, compared to a typical annotation task, the use of physicians is much more expensive and difficult to schedule. Therefore, depending on the complexity/specialization of the clinical task, medical students might be an optimal alternative.

The reliability of an annotated corpus is another important aspect to be aware of. Most of the work relies on the inter-annotator agreement (IAA) calculation as the main metric to assess reliability. There are different methods used to calculate the IAA. Artstein and Poesio (2008)[40] surveyed most of them (e.g., observed agreement, Krippendorff's alpha, Cohen's kappa), and discussed their use in multiple annotation tasks. As pointed out by many researchers [13,37,41,42], Cohen's kappa and other chance correction approaches (which are vastly used in classification tasks) are not the most appropriate measure for named entity annotation, because the probability of agreement by chance between annotators is nearly zero when one labels text spans. In addition to the method used to calculate the IAA, what value is considered good and represents a quality gold standard? Hovy and Lavid (2010)[39] state that the annotator manager needs to determine the acceptable IAA values based on its goals, and when it comes to using the corpus to train ML algorithms, one should aim to have enough data with realistic agreement values considering the desirable task.

In the medical area, a well-known and accepted convention for IAA "strength" values is the one proposed by Landis and Koch (1977)[43], in which 0.41<=IAA<=0.60 is moderate, 0.61<=IAA<=0.80 is substantial, and IAA>= 0.81 is almost perfect.
Artstein and Poesio[40] claim that the adequate level of agreement for specific purposes is obscure, because different levels of agreement may be good for one purpose and bad for another. They discussed Reidsma and Carletta's (2008) work[44], in which the authors approach the reliability thresholds used in Computational Linguistics (CL), where IAA ≥ 0.8 is considered to be good and 0.8 > IAA

> 0.67 is tolerable. The authors also demonstrate that ML algorithms can tolerate data with low-reliability values, and sometimes, 0.8 reliability measures are not synonymous with good performance. In other words, agreement metrics are weak predictors of ML performance. This agrees to an extent with the discourse in Roberts et al. (2009)[5], who stated, *"The IAAs between double annotators that are given do not therefore provide an upper bound on system performance, but an indication of how hard a recognition task is."*

## 3. Data preparation

Our data are obtained from two different data sources: (1) a corpus of 2,094,929 entries from a group of hospitals in Brazil made in the period between 2013 and 2018, and (2) a corpus originating from a University Hospital based on entries made in the period between 2002 and 2007, which counts with 5.617 entries. In the first dataset, each entry has structured data (e.g., gender, medical specialty, entry date) and unstructured data in a free-text format, representing sections of a clinical narrative (e.g., disease history, family history, and main complaint). The data were obtained from the records of approximately 553,952 patients.

Besides the multi-institutional aspect of the corpus, it covers various medical specialties (e.g., cardiology, nephrology, and endocrinology) and clinical narrative types (e.g., discharge summaries, nursing notes, admission notes, and ambulatory notes).

The second dataset has only one document type (i.e., discharge summaries) and comes from the Cardiology Service Center exclusively. The data configuration has structured data (i.e., gender, birth date, begin date, end date, and icd-10 code) and just one free-text data field, concerning the discharge summary. The texts from both datasets have some already known characteristics related to clinical narratives in general [45], such as uncertainty, redundancy (often due to copy and paste), high use of acronyms and medical jargon, misspellings, fragmented sentences, punctuation issues, and incorrect lower and uppercasing. Some text examples are presented in Table 1. The de-identification process is described in the "annotation tool" section.

| Type/Specialty | Original narrative | Translated narrative |
|---|---|---|
| Discharge summary Cardiology | PACIENTE DIABÉTICA, HIPERTENSA, CARDIOPATIA ISQ. COM IMPLANTE DE STENT EM DAE EM JUL/03 INTERNOU COM QUADRO DE ANGINA INSTÁVEL. TRANSFERIDA PARA O SERV DE HEMODINÂMICA, REALIZOU CAT SENDO SUBMETIDA A ACTP EM LESÃO DE ÓSTIO DA SEGUNDA DIAGONAL. PROCEDIMENTO REALIZADO COM SUCESSO ANGIOGRÁFICO. RECEBE ALTA ASSINTOMÁTICA. Paciente ex-tabagista, vem à emergência com quadro de dispnéia progressiva, ortopnéia, dispnéia paroxística noturna, edema de membros inferiores, turgência jugular. Diagnóstico de insuficiência cardíaca, com classe funcional IV (NYHA) na chegada. Sem história de dor torácica. ECG da chegada sem alterações. Marcadores de necrose miocárdica normais. Manejado para insuficiência cardíaca com boa resposta clínica. Ecocardiograma demonstrando dilatação de cavidades (AE = 5,3 cm, DDVE = 7,0, DSVE = 5,8), disfunção sistólica (FEVE = 35%) por hipocinesia difusa, septo e parede posterior de 0,9 cm, insuficiência mitral e tricúspide leves e PSAP = 52 mmHg. Realizado investigação etiológica com sorologia negativa para Chagas, cintilografia demonstrando necrose apical, sem condições de discriminar isquemia. Optado então pela realização de cateterismo cardíaco, que revelou artéria circunflexa dominante e livre de lesões significativas; artéria coronária direita livre com sinais de aterosclerose, mas sem lesões significativas; artéria descendente anterior de pequeno calibre, com lesão de cerca de 60% no terço proximal e lesão crítica no terço médio. Após revisão do filme, observou-se tratar de lesão de difícil manejo percutâneo, devido à sua extensão e ao pequeno calibre da artéria descrita. Após discussão do caso, optou-se por manejo clínico devido ao fato do paciente não apresentar angina, ter respondido com sucesso à terapêutica instituída e não apresentar evidência clara de benefício atual com procedimento de revascularização. Impressão de que a lesão em DAE não explicaria a hipocinesia difusa apresentada pelo paciente, devendo ser portanto doença aterosclerótica coexistindo em um coração com miocardiopatia dilatada. Realizado ainda espirometria que evidenciou distúrbio obstrutivo moderado. DCE estimada em 57 ml/min. Paciente recebe alta em bom estado geral, afebril, eupnéico, em otimização do tratamento para ICC (já em uso de betabloqueador, IECA e espironolactona), com plano de ajustes de doses a nível ambulatorial. OBS: peso na alta: 76 Kg. | DIABETIC PATIENT, HYPERTENSE, ISCHEMICAL CARDIOPATHY. WITH STENT IMPLANT IN LAD IN JUL / 03 HOSPITALIZED WITH SYMPTOMS OF UNSTABLE ANGINA. TRANSFERRED TO THE SERVICE OF HEMODYNAMIC, PERFORMED CATHETERISM, SUBMITTED TO PCTA IN THE SECOND DIAGONAL INJURY. PROCEDURE PERFORMED WITH ANGIOGRAPHIC SUCCESS. ASYMPTOMATIC HOSPITAL DISCHARGE. Ex-smoker patient comes to the emergency room with progressive dyspnea, orthopnea, paroxysmal nocturnal dyspnea, lower limb edema, and jugular turgence. Heart failure diagnosis, with functional class IV (NYHA) upon arrival. No history of chest pain. ECG on arrival without change. Normal myocardial necrosis markers. Managed for heart failure with good clinical response. Echocardiogram showing cavity dilatation (LA = 5.3 cm, LVDD = 7.0, LVSD = 5.8), systolic dysfunction (LVEF = 35%) due to diffuse hypokinesia, a 0.9 cm septum and posterior wall, mild mitral and tricuspid regurgitation, and APSP = 52 mmHg. Etiological investigation with Chagas negative serology, scintigraphy showing apical necrosis, unable to discriminate ischemia. Then opted for catheterization which revealed a dominant circumflex artery free of significant lesions; free right coronary artery with signs of atherosclerosis but no significant lesions; small anterior descending artery with a lesion of about 60% in the proximal third and critical injury in the middle third. After review of the film, it was observed that it was a difficult percutaneous management injury owing to its extension and the small caliber of the described artery. After discussion of the case, we opted for clinical management because the patient did not have angina, successfully responded to the therapy instituted, and did not present clear evidence of being benefitted by the revascularization procedure. The impression that the lesion in LAD would not explain the diffuse hypokinesia presented by the patient; therefore, atherosclerotic disease coexisting in a heart with dilated cardiomyopathy. Accomplished yet spirometry that showed moderate obstructive disorder. DCE estimated at 57 ml / min. Patient is discharged in good general condition, afebrile, eupneic condition, optimizing treatment for CHF (already using beta-blocker, ACEI and spironolactone), with outpatient dose adjustment plan. OBS: weight in the high: 76 Kg |
| Ambulatory note Nephrology | NEFROPATIA DIABETICA EM TTO CONSERVADOR<br>CANDIDATA A TX RENAL PREEMPTIVO<br>LIBERADA PELA URO E ANESTESIO<br>CANDIDATA A TX RENAL PREEMPTIVO<br>ASSINTOMÁTICA, EXCETO PELOS SINAIS E SINTOMAS ASSOCIADOS A NEUROPATIA PERIFERICA ( DIABETICA / UREMIA)<br>SEM SINTOMAS URINARIOS<br>AO EXAME PA 150/100 P 108 T 36 DIURESE FRR NORMAL<br>HIPOCORADA +<br>CPP LIVRES<br>PC RITMO REGULAR, TAQUICARDICO<br>ABD RHA+, PLANO, FLÁCIDO, CIC CX CST<br>MMII PULSOS PRESENTES E SIMETRICOS | DIABETIC NEPHROPATHY IN CONSERVATIVE TREATMENT<br>Preemptive Kidney Transplant Candidate<br>RELEASED BY UROLOGY AND ANESTHESIOLOGY<br>Preemptive Kidney Transplant Candidate<br>ASYMPTOMATIC, EXCEPT FOR SIGNS AND SYMPTOMS ASSOCIATED WITH PERIPHERAL NEUROPATHY (DIABETIC / UREMIA)<br>NO URINARY SYMPTOMS<br>ON EXAMINATION BP 150/100 HR 108 T 36 DIURESE RR NORMAL<br>PALLOR +<br>FREE LF<br>CS REGULAR Rhythm, Tachycardic<br>ABDOMEN RHA +, FLAT, FLAT, CIC CX CST<br>LLLL PRESENT AND SYMMETRICAL PULSES |
| Nursing note<br>Not defined | Pcte com RNM de crânio agendada para hoje às 23:00h. Por volta das 21:00h pcte apresentou quadro de confusão mental, seguida de crise convulsiva generalizada, prontamente atendido na sala de poli, com MCC + oximetria digital de pulso + PNI contínuos. Instalado O2, medicado CPM e mantido em observação no leito. Hidantalizado pela R1 Vital Brasil da neurocirurgia, procedimento realizado sem intercorrências. Pcte bastante sonolento, mantido em sala de poli e suspenso RNM por hora. Diurese espontânea, com controle através de uropen. SSVV às 05:45h PA = 133/74mmhg, FC = 114bpm, SpO2 = 93%. Conforme orientação da neurocirurgia, mantém observação na sala de poli sob cuidados intensivos de enfermagem. CHOQUE NAO ESPECIFICADO | Patient Skull MRI scheduled today at 23:00. At around 21:00, the patient presented with mental confusion, followed by generalized seizure, promptly treated in the multiple trauma room, with MCC + digital pulse oximetry + continuous NIBP. Installed O2, medicated as prescribed and kept under observation in bed. Hidrantalized by R1 Vital Brasil of neurosurgery, procedure performed without complications. Very sleepy patient kept in emergency room and suspended MRI for hour. Spontaneous diuresis, with uropen control. VVSS at 05:45 h BP = 133 / 74 mmhg, HR = 114 bpm, PsO2 = 93%. As directed by neurosurgery, maintains observation in the emergency room under intensive nursing care. SHOCK NOT SPECIFIED |

**Table 1. Samples of different types of clinical narratives from our corpus.**

## 3.1 Document selection

The original and main focus of the intended semantic annotation was two-fold: (i) to support the development of a NER algorithm to be used in a summarization method and (ii) to evaluate a semantic search algorithm, both focusing on cardiology and nephrology specialties. Thus, we selected almost 500 clinical notes from both medical specialties (including two patients' complete longitudinal records). Owing to the lack of corpora for pt-br, we decided to increase the scope of study to support other bio-NLP tasks and medical specialties. We randomly selected documents from other medical areas to complete 1,000 clinical narratives, assuming that the data are satisfactorily consistent and representative to train an ML model. Table 2 shows the number of documents per specialty. The average character token size was ~148 and the average sentence size was approximately 10 tokens.

| Specialty | Number |
|---|---|
| Cardiology | 260 |
| Nephrology | 157 |
| Orthopedy | 126 |
| Not defined | 122 |
| Surgery (general) | 61 |
| Neurology | 45 |
| Neurosurgery | 32 |
| Dermatology | 23 |
| Ophthalmology | 22 |
| Endocrinology | 19 |
| Gastroenterology | 16 |
| Otolaryngology | 14 |
| Pneumology | 11 |
| Others | 92 |

**Table 2**. Medical specialties frequency table

Note that several documents are assigned as "Not defined," because this is one of the majority classes in the corpus we received. Nevertheless, by looking at these documents, we can conclude that these patients are (a) under the care of multiple medical specialties (e.g., patient with multiple trauma, in ICU) or (b) in the middle of a diagnostic investigation. Moreover, we grouped the specialties with less than 10 documents as "Others" (e.g., urology, oncology, gynecology, rheumatology, proctology).

## 3.2 Text organization

In Table 3, we present the available data in each entry of the database (concerning our first and main data source). To have a unique text file per entry, we concatenated all free-text fields into a single text document to be annotated.

Besides the already known issues in clinical text, our database has other issues. The medical staff is supposed to write the patient's clinical note in all the free-text fields. The EHR application has one textbox for each field, and these sections serve as the clinical narrative sections. However, as most of the clinicians enter all the text in the history-of-disease field only, with the others remaining empty, making it difficult to search for specific information in the narrative (e.g., look for family history). Additionally, some text is written completely in upper case letters and interfering directly in some text processing, such as finding abbreviations and identifying proper nouns.

| Field | Data type |
|---:|---|
| occurrence-id | Number |
| patient-id | Number |
| gender | Text |
| birth-date | Date |
| inclusion-date | Date |
| discharge-date | Date |
| discharge-type | Text |
| discharge-reason | Text |
| icd-10 | Text |
| medical-specialty | Text |
| care-reason | Text |
| main-complaint | Free-Text |
| history-of-disease | Free-Text |
| past-history | Free-Text |
| family-history | Free-Text |
| physical-examination | Free-Text |
| main-diagnosis-hypothesis | Free-Text |
| initial-plan | Free-Text |
| observations | Free-Text |

**Table 3.** Database entry data configuration

## 4. Annotation schema

In this section, we describe the entire annotation schema, including the conception and evolution of the annotation guidelines, the development of a tool to support and improve the annotation workflow, and finally an overview of the annotation process and its experimental setup. The steps that were followed consider the lessons learned from other similar annotation projects reviewed in section 2.

### 4.1 Annotation guidelines

To ensure the quality of a gold standard, it is crucial to maintain the homogeneity of annotation during the entire process. To provide guidance to annotators and answer their possible questions, we defined a set of guidelines, in which we explained, in detail, how to annotate each type of concept and showed examples of what should be annotated and what should not.

The first step was to define which information we wanted to annotate within the text. Regarding the clinical concepts, we opted to use the UMLS semantic types[6] (STY) as our annotation tags (e.g., "Body Location or Region," "Clinical Attribute," "Diagnostic Procedure," "Disease or Syndrome," "Finding," "Laboratory or Test Result," "Sign or Symptom," and "Therapeutic or Preventive Procedure"). Table 4 presents some of the most used STYs with examples.

We decided to use the UMLS STYs because we needed a high granularity annotation, as we wanted a ground-truth to evaluate a semantic search algorithm that labels entities using the STYs (more than a hundred types). We assumed the risk of agreement loss (owing to the reliability and granularity trade-off discussed in section 2), but, at the same time, with greater coverage of the concepts in the corpus, the gold standard could be utilized in a higher number of bio-NLP tasks. Even when the task has a low granularity approach, it is possible to export the actual annotations to their corresponding UMLS semantic groups[7] (SGR). The second reason for our use of the UMLS STYs relies on the UMLS Metathesaurus resource, which can serve as an important guide to annotators, as they can search for a specific concept to make sure of the STY they are annotating.

---

[6] https://www.nlm.nih.gov/research/umls/META3_current_semantic_types.html
[7] https://metamap.nlm.nih.gov/SemanticTypesAndGroups.shtml

| SGR | STY | Original examples | Translated examples |
|---|---|---|---|
| Anatomy | Body Location or Region | MEIA TALA GESSADA EM <u>MIE</u><br>apresenta edema em <u>região craniana</u><br><u>ABDÔMEN</u> PLANO E FLÁCIDO | Half-length plaster cast in <u>LLL</u><br>presents edema in the <u>cranial region</u><br>FLAT AND FLACID <u>ABDOMEN</u> |
| Anatomy | Body Part, Organ, or Organ Component | acesso venoso central em <u>jugular D</u><br>ACESSO VENOSO PERIFERICO EM <u>BRAÇO DIREITO</u> | <u>right jugular</u> central venous access<br><u>RIGHT ARM</u> PERIPHERAL VENOSOUS ACCESS |
| Chemicals & Drugs | Organic Chemical | Fez uso de <u>atenolol</u> por 3 anos<br>cefaléia em regiao parietal bilateral que melhora com <u>dipirona</u> | used <u>atenolol</u> for 3 years<br>headache in bilateral parietal region improved with <u>dipyrone</u> |
| Chemicals & Drugs | Pharmacologic Substance | asmatica em uso de <u>salbutamol</u> e <u>budesonida</u> | asthmatic person using <u>salbutamol</u> and <u>budesonide</u> |
| Concepts & Ideas | Temporal Concept | <u>POI</u> DE LAVAGEM + CURETA DE TECIDO NECRÓTICO<br>Paciente em <u>Pré-operatório</u> de FX fêmur | WASHING <u>IP</u> + NECROTIC TISSUE CURETAGE<br><u>Preoperative</u> patient of femur fracture |
| Devices | Drug Delivery Device | cloreto de potassio a 42 ml/h em <u>bomba de infusão</u> | potassium chloride at 42 ml/h in <u>infusion pump</u> |
| Devices | Medical Device | <u>AVP</u> em MSE com soroterapia em curso<br><u>SVD</u> com diurese efetiva | <u>PVA</u> in LUL with ongoing serotherapy<br><u>DBP</u> with effective diuresis |
| Disorders | Disease or Syndrome | REFERE <u>HIPERTENSÃO</u> E <u>DIABETES</u> EM USO DE INSULINA.<br><u>SINDROME DE GUILLAIN BARRE</u>. | REFERS <u>HYPERTENSION</u> AND <u>DIABETES</u> IN INSULIN USE<br><u>GUILLAIN BARRE SYNDROME</u> |
| Disorders | Finding | RETORNOU DO CC <u>LÚCIDO</u>, <u>ORIENTADO</u>, <u>COMUNICATIVO</u><br><u>consciente</u>, <u>comunicativo</u>, pupilas <u>isocóricas fotoreagentes</u> | RETURNED <u>LUCID</u> FROM SC <u>CONSCIOUS</u>, <u>COMMUNICATIVE</u><br><u>conscious</u>, <u>communicative</u>, <u>photoreagent isochoric</u> pupils |
| Disorders | Injury or Poisoning | <u>TRAUMA CRÂNIOCERVICAL</u> APÓS QUEDA<br><u>FRATURAS MULTIPLAS</u> DA COLUNA TORACICA. | <u>SKULL-CERVICAL TRAUMA</u> AFTER FALL<br><u>MULTIPLE FRACTURES</u> IN THORACIC COLUMN |
| Disorders | Sign or Symptom | relata <u>cefaléia</u><br>SINAIS VITAIS ESTAVÉIS, REFERE <u>ALGIA</u> | reports <u>headache</u><br>STABLE VITAL SIGNS, REFERS <u>PAIN</u> |
| Living Beings | Patient or Disabled Group | <u>paciente</u> nega queixas, nega dor, dispnéia.<br><u>Pcte</u> com cultura de Secreção Tibial | <u>patient</u> denies complaints, denies pain, dyspnea.<br><u>Ptt</u> with Tibial Secretion culture |
| Living Beings | Professional or Occupational Group | Orientada a <u>equipe de enfermagem</u> que o mesmo esta em jejum<br>segundo a <u>farmacêutica</u> e o <u>médico</u> | Advised <u>nursing staff</u> that the patient is fasting<br>according to the <u>pharmacist</u> and the <u>doctor</u> |
| Organizations | Health Care Related Organization | CONFORME ROTINA DA <u>UTI</u><br>RETORNOU DO <u>CC</u> ÀS 14:30HRS | AS <u>ICU</u> ROUTINE<br>RETURNED FROM <u>SC</u> AT 2:30pm |
| Phenomena | Laboratory or Test Result | <u>Glicose 335</u>; <u>LDH 223</u>;<br><u>Teste rápido para HIV negativo</u> | <u>Glucose 335</u>; <u>LDH 223</u>;<br><u>HIV negative rapid test</u> |
| Physiology | Clinical Attribute | <u>PA</u> = 130/70<br><u>PESO</u> 67,4 | <u>BP</u> = 130/70<br><u>WEIGHT</u> 67.4 |
| Procedures | Diagnostic Procedure | <u>AUSCULTA PULMONAR</u>; MV +, RONCOS DIFUSOS EM BASES<br>Monitorização cardíaca contínua, <u>PAM</u> e <u>oximetria digital</u>. | <u>PULMONARY AUSCULTATION</u>; VM +, DIFFUSED WHEEZES IN BASES<br>Continuous cardiac monitoring, <u>MAP</u>, and digital oximetry. |
| Procedures | Health Care Activity | EM <u>ACOMPANHAMENTO</u> NA ENDOCRINO DO HC<br><u>Internamento</u> em janeiro por taquicardia atrial com aberrância | <u>FOLLOW-UP</u> ON ENDOCRINOLOGY AT HC<br><u>Admission</u> in January for aberrant atrial tachycardia |
| Procedures | Therapeutic or Preventive Procedure | <u>SVD</u> COM 100 ML DEBITO SEM GRUMOS<br>IRC EM <u>DIALISE</u> | <u>DC</u> WITH 100 ML DEBIT WITHOUT GROUNDS<br>CRF IN <u>DIALYSIS</u> |
| N/A | Abbreviation | CONFORME ROTINA DA <u>UTI</u>[Unidade de Terapia Intensiva]<br>MEIA TALA GESSADA EM <u>MIE</u>[Membro inferior esquerdo] | AS <u>ICU</u>[Intensive Care Unit] ROUTINE<br>Half-length plaster cast in <u>LLL</u> [Lower left limb] |
| N/A | Negation | Paciente eupnéico e <u>afebril</u><br>Paciente <u>nega</u> algia<br><u>SEM</u> IRRADIAÇAO | Eupneic and <u>feverless</u> patient<br>Patient <u>denies</u> pain<br><u>NO</u> IRRADIATION |

**Table 4.** Text samples containing the most used STYs. The underlined passages indicate the annotated concepts.

Moreover, the UMLS REST API[8] allows our annotation tool to automatically suggest the STY for some clinical concepts. Furthermore, as two important bio-NLP tasks are not covered by the STYs, we added two more types in our tagset, the "Negation" and "Abbreviation" tags. The first one aims to identify negation cues associated with clinical concepts (already tested in a negation detection algorithm presented in a later section). The "Abbreviation" type was incorporated to help us in the process of abbreviation disambiguation. Sometimes, when we want to extract semantic meaning from clinical text, the semantic value of a concept alone is not enough to infer important events and situations. Hence, we incorporated the annotation of relations between clinical concepts to the guidelines. The relation annotation schema was partially derived from the UMLS Relationships Hierarchy. Unlike the concept schema, we decided to use a restricted set of tags; therefore, instead of 54 UMLS relationship types, we used only one to simplify the relation annotation (as it was not our main focus). The RTYs included only the "*associated_with*" and "*negation_of*" (not a UMLS RTY; it was added to complement the Negation STY) RTYs. There are five major non-hierarchical RTYs (i.e., *conceptually_related_to, functionally_related_to, physically_related_to, spatially_related_to, temporally_related_to*) that connect concepts by their semantic values. We represent them by using their parent RTYs only, the "*associated_with*" RTY. Depending on the chosen STY, it is possible to infer the sub-types of "*associated_with*" automatically. Once the concepts and relationships were defined, an annotation script was established, in which the annotator should first label all the concepts and then annotate the relations. We followed this order because Campillos et al. (2018)[4] found that the agreement between annotators was higher when annotating was performed this way.

Deleger et al. (2012)[37] stated that the most difficult STY to annotate was "Finding," because it is a very broad type that can correspond to signs/symptoms (e.g., "fever"), disease/disorders (e.g., "severe asthma"), laboratory or test results (e.g., "abnormal ECG"), and general observations (e.g., "in good health"). To avoid disagreements, we decided to simplify the definition of "Finding." Annotators should always give preference to disease/disorders and lab results STY's over the "Finding" STY. Only results of physical examination considered to be normal should be marked as "Finding" (e.g., "flat abdomen" and "present diuresis"). The abnormal ones should be labeled as "Sign or Symptom". We know that this can cause discrepancies between UMLS concepts and our annotation, but it makes more sense for our task. Using these definitions, we prepared the first draft of the guidelines, which was given to the annotators. Furthermore, we provided training that acquainted them with the annotation tool and allowed them to realize some of the difficulties in the process. Then, an iterative process was started to enhance the guidelines, and check the consistency of annotation between annotators (more details on Inter Annotator Agreement in the Results section), and provide feedback on their work. We assumed that, if in 3 consecutive rounds the agreement stayed stable (no significate reduction or improvement), there was no room for guideline adaptation, and the final annotation process could be initiated. A flowchart of the process is given in Figure 1, and it is similar to what Roberts et al. (2009)[5] and others used.

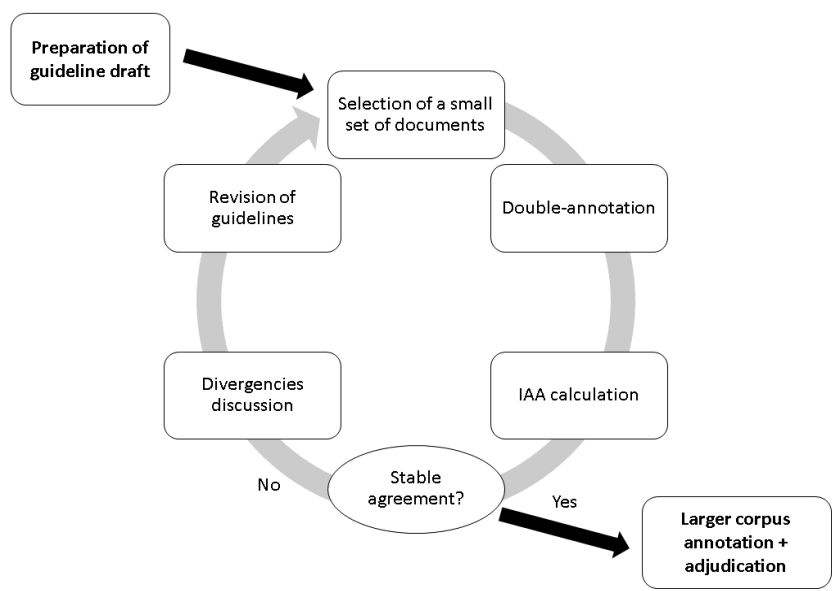

**Figure 1.** Revision and quality verification process of the annotation guidelines

---

[8] https://documentation.uts.nlm.nih.gov/rest/home.html

## 4.2 Annotation tool

The previously discussed issues and difficulties related to clinical annotation indicate that we needed an annotation tool that can ease and accelerate the annotators' work. We analyzed Andrade et al.'s (2016)[46] review on annotation methods and tools applied to clinical texts and decided to build our own tool. This approach ensured that all annotators could share the same annotation environment in real-time and work anywhere/anytime without technical barriers (i.e., web-based application). Furthermore, the project manager could better supervise and organize all the work and assign the remaining work to the persons involved. Moreover, as we used UMLS semantic types in our scheme, it would be desirable to use the UMLS API and other local resources (e.g., clinical dictionaries) supporting the text annotation by making annotation suggestions to the user without pre-annotating it. Finally, we needed a tool that fitted exactly into our annotation workflow, with the raw data input into our environment, and a gold standard output at the end of the process, dispensing the use of external applications. Our tool workflow was composed of six main modules:

- Importation: import data files into the system
- Review: manually remove PHI information that the anonymization algorithm failed to catch
- Assignment: allocate text to annotators
- Annotation: allow labeling of the clinical concepts within the text with one or multiple semantic types, supported by the Annotation Assistant feature
- Adjudication: resolve double-annotation divergences and creation of the gold standard
- Exportation: exports the gold standard as JSON or XML

The aforementioned Annotation Assistant component was developed to prevent annotators from labeling all the text from scratch by giving them suggestions of possible annotations based on (a) previously made annotations and (b) UMLS API exact-match and minor edit-distance lookup. Further details on technical aspects, modules functionalities, and experiments showing how the tool affects the annotation time and performance are reported on [47].

## 4.3 Annotation process

Besides the advice and recommendations found in section 2, similar to Roberts et al. (2009)[5], we decided to follow a well-known annotation methodology standard [48]. Furthermore, we added a guidelines agreement step, in which all the text was double-annotated with the differences resolved by a third experienced annotator (i.e., adjudicator), and documents with low agreement were not included in the gold standard. Pairing annotators to perform a double annotation of a document prevents bias caused by possible mannerisms and recurrent errors of a single annotator. Moreover, it is possible to check the annotation quality by measuring the agreement between both annotators.

It is almost impossible to achieve an absolute truth in such an intricate annotation effort like this one. To reach as close as possible to a consistent ground-truth, we adopted the use of an adjudicator, who was responsible for resolving the differences between the annotators. It is worth mentioning that the adjudicator cannot remove annotations made by both annotators, and neither create new annotations, hampering a gold standard made with the opinion of a single person. After the guideline maturing process, we started the final development stage of the gold standard, which posteriorly was divided into ground-truth phases 1 and 2. We decided to recruit annotators with different profiles and levels of expertise to give us different points of view during the guideline definition process, and to determine if there were differences in annotation performance between annotators with different profiles.

**Ground-truth phase 1** counted with a team of three persons: (1) a physician with experience in documenting patient care and participation in a previous clinical text annotation project; (2) an experienced nurse; and (3) a medical student who already had ambulatory and EHR's use experience. The nurse and the medical student were responsible for the double-annotation of the text, and the physician was responsible for adjudicating them. When the process was almost 50% complete (with 496 documents annotated and adjudicated), we managed to engage more people to assist in finishing the task (in what we called **ground-truth phase 2**). An extra team of 6 medical students, with the same background as the first one, were recruited (Figure 2 illustrates these phases). We held a meeting in which we presented the actual guidelines document and trained them on using the annotation tool.

In phase 2, we had two adjudicators, the physician, and the nurse. We added the nurse as an adjudicator as we needed one extra adjudicator during this phase, and the nurse had more hospital experience than medical student 1. Then, we had a homogeneous group of seven medical students annotating the texts. The physician, the nurse, and medical student 1 supervised the first set of annotations of all the students. The number of documents to be annotated were divided equally between the annotators and adjudicators, and the selection of double-annotators for each document was made randomly, as was done for the adjudicators. It is worth noting that besides the people mentioned above, who worked directly with annotation and adjudication, we had a team of Health Informatics researchers who participated in supporting the annotation project with other activities, including annotation tool development, guidelines discussion, and annotation agreement feedback.

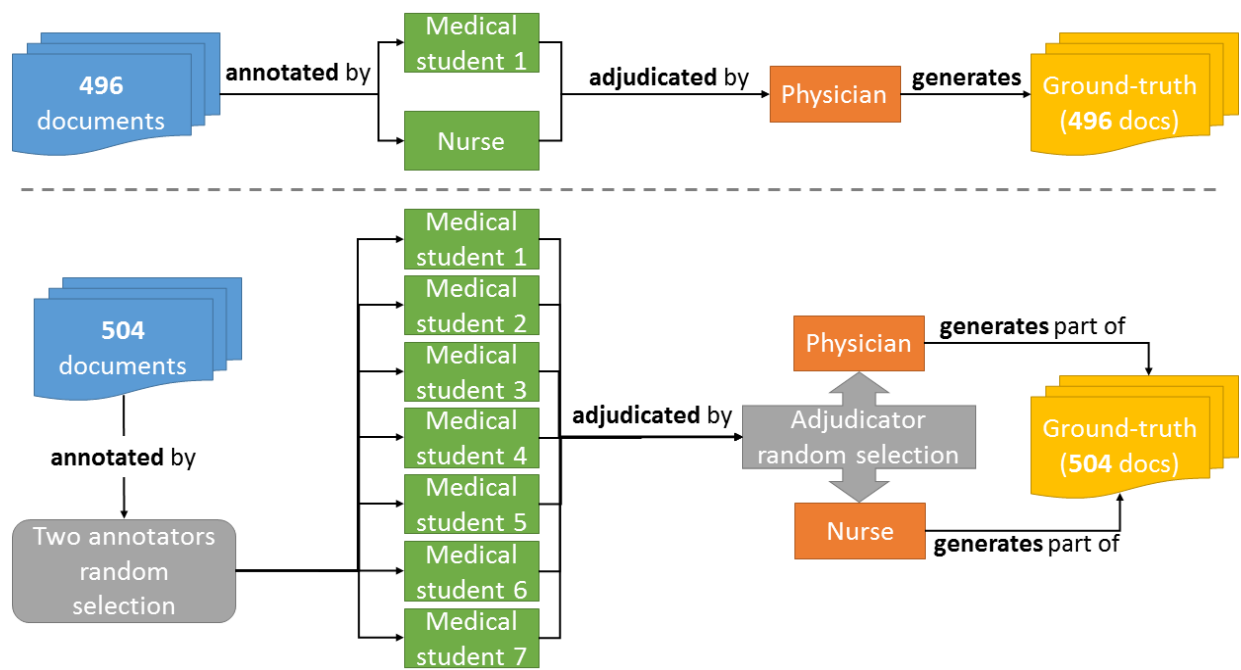

**Figure 2.** Annotation process overview, including ground-truth phases 1 and 2, which are located above and below the dashed line, respectively.

### 4.4 Corpus reliability and segmentation

Taking advantage of the fact that we had double-annotated the entire collection of documents, we calculated the IAA of all the data using the *observed agreement* metric, as presented in the following equation (no need for chance-correction calculations, as described in section 2). For the *strict* version of IAA, a situation was considered a match, when the two annotators label the same textual span with an equal semantic type. All other situations were calculated as a non-match. We reported the *lenient* version of IAA as well, which considers partial matches, that is, the annotations that have overlaps in the selected textual spans (with the same STY); these are counted as a half-match in the formula. The third version of IAA, called *flexible*, was calculated. We transformed the annotated STY to its corresponding SGR (e.g., "Sign or Symptom," "Finding," and "Disease or Syndrome" STYs are converted to the "Disorder" SGR). Then, we performed a comparison to determine whether the SGRs are equal (the textual span needs to be the same). Finally, the fourth version of IAA was *relaxed*, i.e., we considered partial textual spans (overlaps) and SGRs at the same time.

$$IAA = \frac{matches}{matches + non\_matches}$$

To isolate the concept agreement scores from the relationship score, we reported the relationship IAA values by considering only those relationships in which both annotators labeled two of the connected concepts. Otherwise, if an annotator did not find one of the concepts involved, the relationship IAA would be directly penalized.

Boisen et al. (2000)[48] recommend that only documents with an acceptable level of agreement should be included in the gold standard, and we followed their recommendation. However, because of the scarcity of this kind of data in pt-br bio-NLP research, and as the limited amount of annotated data is often a bottleneck in ML [49], we did not exclude documents from our corpus, but opted to segment it in two, namely **gold** and **platinum**. This division was made based on the IAA values of each annotated document, where documents with an **IAA greater than 0.67** belong to the gold standard and all the other ones to platinum. We picked the 0.67 threshold because is the one Artstein and Poesio (2008)[40] discussed to be a tolerable value. Additionally, we think the 0.8 threshold is rigorous considering the complexity of our task and the number of persons involved in it. The task complexity is explained by the heterogeneity of the data, which are obtained from multiple institutions, various medical specialties, and different types of clinical narratives. The study that comes closest in data diverseness to ours is Patel et al. (2018)[30], with the exception that their data come from a single institution. Moreover, despite the large amount of data they used, there are differences between their study and ours; for example, they used a coarse-grained annotation scheme by grouping the STYs, which made the labeling less prone to errors. Moreover, we believe that a great portion of errors that caused disagreements came from repeated mistakes on the part of one annotator in the pair, and owing to this, the error could be easily corrected by the adjudicator, as the examples in the following sections reveal.

## 5. Annotation results and analysis

This section compiles quantitative and qualitative results regarding our corpus development. We detail the IAA information used to segment the corpus and analyze the errors found during the annotation. Finally, we introduce and present the results of two bio-NLP applications that had already used the current corpus in their development.

### 5.1 Corpus statistics

The corpus development involved 8 annotators, 2 adjudicators, and 4 Health Informatics researchers, totaling a team of 14 people. Our corpus comprehended 100 UMLS semantic types representing the entities, 2 extra semantic types typifying Abbreviations and Negations, and 2 relationship types defining the relations between clinical entities. The annotation process was 100% double-annotated and adjudicated, and lasted 14 months, resulting in a corpus composed of 1,000 documents (148,033 tokens), with 65,129 entities and 11,263 relations labeled. In Table 5, we present the corpus size, considering the gold/platinum divisions. Tables 6 and 7 show the number of annotations per STY and RTY.

### 5.2 Inter annotator agreement

We calculated the average agreement between all the 1,000 double-annotated documents in the corpus using four different IAA versions for the concepts (i.e., strict, lenient, flexible, and relaxed) and the regular version for relations. We achieved an average *strict* IAA of **~0.71** and **~0.92** for the *relaxed* version in the concept annotation task. For the relations, the IAA was **~0.86**. In Table 8 we detail the average IAA values for the entire corpus, and Figure 3 presents the average agreement considering the most frequent STYs. Table 9 shows the IAA per RTY.

The results reveal that even with a complete annotation environment, composed by a refined set of guidelines, use of a personalized annotation tool, clinically trained annotators, and constant reliability analysis, it is extremely difficult to reach a perfect agreement. Overall, we believe that we built a good quality corpus, with IAA values comparable to other clinical semantic annotation studies if we consider the specificities of it (described later in this section). It is worth noting that some other studies evaluated their corpus by calculating the agreement between annotators and adjudicators, which typically produces superior agreement numbers when compared to IAA. This event can be seen in Bethard et al. (2016)[27], who achieved a 0.73 IAA and 0.83 annotator-adjudicator agreement. Another important detail is that we double-annotated 100% of our documents, not just portions of the corpus like most of the related work, which affects the final results as below average agreement values appear during certain project phases (e.g., because of a new annotation team or guidelines changes), and makes us believe that we have a clear and trustworthy view of the homogeneity of our corpus.

| Segment | Documents | Entities | Relations |
|---|---|---|---|
| Gold | 613 | 41,588 | 7,344 |
| Platinum | 387 | 23,541 | 3,919 |
| **TOTAL** | **1,000** | **65,129** | **11,263** |

**Table 5.** Corpus size considering gold and platinum divisions

The IAA scores by STYs (shown in Figure 3) corroborate with other authors regarding the difficulty difference between entity types. For example, "Disease or Syndrome" strict IAA was ~0.67 and "Pharmacologic Substance" was ~0.88, probably because the first one is composed mainly of multi-word expressions and the second one of single tokens. The agreement calculation used "less exact" approaches (considering SGRs over STYs) because there was a need to compare our results with those of other clinical semantic annotation studies that grouped the label categories in a few coarse-grained types, like the MERLOT, MiPACQ and MedAlert corpora [4,32,34]. Our approach allows the annotator to use all the semantic types of UMLS, which are more error-prone, particularly if we consider STYs in the same branch of the UMLS hierarchy tree (e.g., "Sign or Symptom" and "Finding"). We want to emphasize that, besides the granularity issue, our corpus development faced other challenges regarding its complexity. Using documents from multiple institutions brings to light some extra difficulties, like dealing with different text formats due to specific institutional workflows and new sets of local abbreviations/acronyms to be aware of. To the best of our knowledge, no other clinical annotation study has covered documents from multiple institutions.

| SGR | STY | Entities |
|---|---|---:|
| Anatomy | Body Location or Region | 1,452 |
| Anatomy | Body Part, Organ, or Organ Component | 1,373 |
| Chemicals & Drugs | Organic Chemical | 2,000 |
| Chemicals & Drugs | Pharmacologic Substance | 3,013 |
| Concepts & Ideas | Quantitative Concept | 3,953 |
| Concepts & Ideas | Qualitative Concept | 500 |
| Concepts & Ideas | Temporal Concept | 1,663 |
| Devices | Medical Device | 1,617 |
| Disorders | Disease or Syndrome | 2,650 |
| Disorders | Finding | 6,867 |
| Disorders | Injury or Poisoning | 521 |
| Disorders | Sign or Symptom | 4,707 |
| Living Beings | Patient or Disabled Group | 844 |
| Living Beings | Professional or Occupational Group | 720 |
| Organizations | Health Care Related Organization | 639 |
| Phenomena | Laboratory or Test Result | 3,079 |
| Physiology | Clinical Attribute | 1,128 |
| Procedures | Diagnostic Procedure | 2,012 |
| Procedures | Health Care Activity | 2,763 |
| Procedures | Therapeutic or Preventive Procedure | 4,791 |
| N/A | Abbreviation | 12,629 |
| N/A | Negation | 2,676 |

**Table 6.** Number of annotations per STY for the entire corpus - considering the most frequent STYs

| RTY | Relations |
|---|---:|
| associated_with | 9,693 |
| negation_of | 1,570 |

**Table 7.** Number of annotations per RTY for the entire corpus

| IAA type | IAA |
|---|---:|
| Strict (full span + STY match) | 0.708 |
| Lenient (partial span + STY match) | 0.834 |
| Flexible (full span + SGR match) | 0.774 |
| Relaxed (partial span + SGR match) | 0.921 |

**Table 8.** Average IAA values for the entire corpus

Handling documents from multiple medical specialties (e.g., Endocrinology, Dermatology) makes the annotators' training and abstraction significantly more difficult. Because with every new document they annotate, the chance to find new challenging, ambiguous, and exception cases is higher than if they tackle documents covering topics in a single medical specialty. The CLEF corpus[5] scope, for instance, covered only narratives from patients diagnosed with neoplasms. The document type diversity could also influence the annotation process, as the documents are produced in different moments during the care workflow and are written by distinct medical professionals (physician, nurse, medical student, or intern); this can sometimes cause interpretation problems owing to their different perspectives.

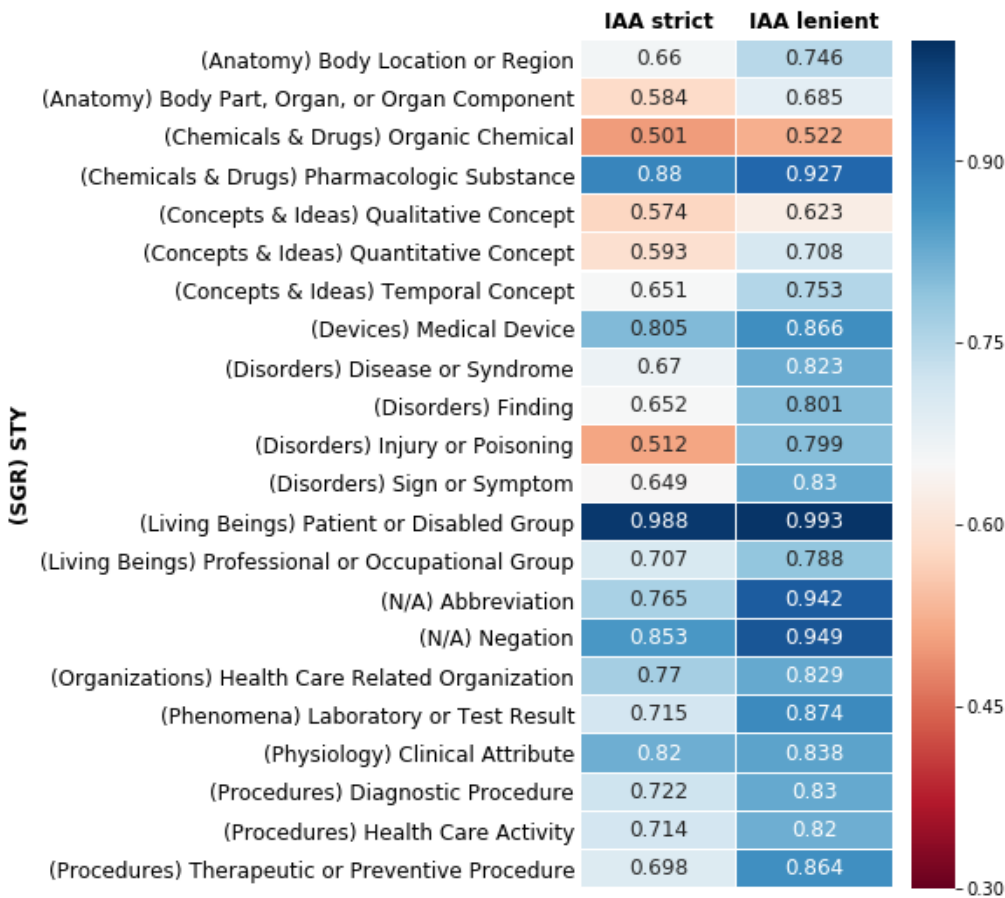

**Figure 3.** Average IAA values for the most frequent STYs

| RTY | IAA |
|---|---|
| associated_with | 0.823 |
| negation_of | 0.914 |

**Table 9.** Average IAA values per RTY

Taking into account all those challenges and particularities of our corpus, we compared our results with previous initiatives and compiled the IAA values of entity and relation annotation of each corpus (see Table 10). The IAA percentage difference for entity annotation are ranging from 2.8% and 18.3% using the strict match, and 3.6% to 23.2% for lenient match. Except for MiPACQ, all the other corpora had better IAA values in strict match, probably due to the issues mentioned previously in the section. However, when we compare the lenient match scores, our performance is better than all other corpora except IxaMed-GS that had the best results of all of them. This led us to believe that our annotators had more trouble in defining the correct text spans than the ones in other projects because when we use a partial span match approach our results improved by 16.9% (from 0.71 to 0.83), and the other corpora improvement ranged from 3.8% and 8.6%. We are not sure if the cause of this is the lack of proper guideline definition, annotators experience or even the document types we used (examples of annotation span issues are detailed in the next section).

If we use the flexible and relaxed match instead of strict and lenient for entity annotation, our result goes from 0.71 to 0.77, and 0.83 to 0.92 respectively, and in that setting, we think we have a fairer evaluation, because the corpus granularity and complexity is more similar to the other works, and then, our results comes closest to the others. Compared to CLEF corpus we achieved the same

IAA for strict vs flexible and increased by 13% the results for lenient vs relaxed. MERLOT and MedAlert had slightly better results (2.6% and 3.9% respectively). IxaMed-GS still have a 9.1% advantage for strict vs flexible, maybe because it is the most specific and least in-depth corpus compared to others, but even so, it has a 2.2% disadvantage for lenient vs relaxed. And finally, we exceeded MiPACQ's results by 10.4% and 18.5% using flexible and relaxed approaches.

For relation annotation, the scenario is completely different from entity annotation, as we used a simpler set of relations compared to other corpora (except IxaMed-GS that used only two relation categories as we did), and probably because of that, we achieved the better results for relation annotation, with the percentage difference ranging from 4.6% to 23.2%.

| Corpus | Type | Strict | Lenient | Flexible | Relaxed |
|---|---|---|---|---|---|
| **CLEF** [5] | entities | 0.77 (8.5%) | 0.80 (-3.6%) | 0.77 (0%) | 0.80 (-13.0%) |
| | relations | - | 0.75 (-12.7%) | - | - |
| **IxaMed-GS** [13] | entities | 0.84 (18.3%) | 0.90 (8.4%) | 0.84 (9.1%) | 0.90 (-2.2%) |
| | relations | - | 0.82 (-4.6%) | - | - |
| **MERLOT** [4] | entities | 0.79 (11.2%) | - | 0.79 (2.6%) | - |
| | relations | - | 0.78 (-9.3%) | - | - |
| **MedAlert** [34,50] | entities | 0.80 (12.6%) | - | 0.80 (3.9%) | - |
| | relations | - | 0.66 (-23.2%) | - | - |
| **MiPACQ** [32] | entities | 0.69 (-2.8%) | 0.75 (-9.6%) | 0.69 (-10.4%) | 0.75 (-18.5%) |
| | relations | - | - | - | - |

Table 10. Comparison between similar clinical annotation projects. In parentheses, the percentage difference in performance compared to our corpus. Note that the IAA values for Flexible and Relaxed match are a copy of Strict and Lenient scores because the other authors did not calculate these metrics and we wanted to know the percentage difference between their values and ours.

**5.3 Error analysis**

The error (or disagreement) analysis showed the most common errors that have impacted the agreement results, and it was performed by the Health Informatics team continuously during the annotation process so that the annotators would be given feedback on their work. As performing a full error analysis for the entire corpus would be highly time-consuming, we only analyzed the part of the documents in which agreement had not reached the 0.67 IAA threshold. Moreover, the adjudicators were already aware of the persistent errors. As expected, a large number of errors occurred at the beginning of the annotation phases (i.e., ground-truth phases 1 and 2), because despite the training, the annotators were still getting used to the annotation process and using the guidelines document. Another common aspect of most of the disagreements is that they are not conceptual, that is, the disagreement does not originate from the semantic value given to the clinical entity, but rather from the different word span selection (term boundaries) normally associated with omission or inclusion of non-essential modifiers and verbs to a term (e.g., "*o tratamento*" vs "*tratamento*" labeled as "Therapeutic or Preventive Procedure" – "*the treatment*" vs "*treatment*").

The STYs high granularity caused two types of annotation divergences. The first one was with regard to the annotation using different STYs with close semantic meaning because they are directly related in the UMLS hierarchy. One of the most occurring errors of this type is related to "Finding" and "Sign or Symptom", even with the simplification that we stated in our Guidelines that says: annotators should always give preference to disease/disorders and lab results STY's over the "Finding" STY. Only results of physical examination considered to be normal should be marked as "Finding". The abnormal ones should be labeled as "Sign or Symptom". Another example of this kind of error is when the annotators should decide between "Medical Device" and "Drug Delivery Device" like with the "*infusion pump*" device. The second type of error associated with the high granularity occurred because some uncommon concepts could be labeled with some infrequent STYs not remembered by the other annotator (e.g., "Element, Ion, or Isotope", "Age Group", "Machine Activity").

Erroneous decomposition of multiword expressions occurred even with many examples explicitly described in the guidelines. This error occurred when one annotator thought a compound term should be labeled as a single annotation and the other annotator as two or more different terms (annotations). There was no unique rule to follow in this case, as it depends on the context. Perhaps this is the reason for this type of error. For instance, the term "*Acesso venoso central direito*" ("*right central venous access*") needs to be decomposed as "*right*" (Spatial concept) and "*central venous access*" (Medical Device), but some annotators simply annotated all the term as "Medical Device". And other terms do not need to be decomposed as "*DRC estágio V*" ("*Chronic kidney disease stage 5*") that must be annotated as "Disease or Syndrome".

We found some errors caused by the ambiguity of certain words that could misinterpreted in its sense, this happened mainly in abbreviations. For instance, "*AC*" could be "*ausculta cardíaca*", "*anticorpo*" or "*ácido*" – "*cardiac auscultation*", "*antibody*" or "*acid*"). The term "*EM*" that could be "*Enfarte do miocárdio*", "*Esclerose múltipla*" or "*Estenose mitral*" – "*Myocardial infarction*", "*Multiple sclerosis*" or "*Mitral stenosis*". We found simple omission errors of some concepts during the analysis as well.

In summary, the STYs performance (Figure 3) reflects the complexity of each STY, for example, the "Pharmacologic Substance" is composed mainly of single-word terms, and "Patient or Disable Group" has just a few terms encompassed by it, explaining their high IAA scores. Unlike "Finding" and "Sign or Symptom" for instance, that have a high frequency and very similar interpretations.

### 5.4 Bio-NLP tasks application

The functionality of an annotated corpus can be tested by applying it in a downstream NLP task. This section provides a brief overview of two bio-NLP studies that already used the corpus presented in this work to train an ML algorithm. The main objective is to prove the consistency and usefulness of our corpus as a rich resource for pt-br clinical tasks and not to present a state-of-the-art algorithm.

*5.4.1 Negation detection*

One constant subject in bio-NLP research is the negation detection, which is often a prerequisite in information extraction tasks because of its important role in biomedical text (e.g., defining the presence or absence of a disease for a patient). Dalloux et al. (accepted/in press manuscript)[51] proposed a cross-domain and cross-lingual negation and scope detection method, in which they used a supervised learning approach supported by a BiLSTM-CRF model with a pre-trained set of Word Embeddings. To train and assess their method in the pt-br clinical scope, they used a segment of our corpus with the negation-related annotations. This includes not only the negation cue labeled with the "Negation" STY, but the concepts related to it using the relation "Negation_of" so that detecting the negation scope would be possible. They achieved a 96.22 F1 score for negation cue detection. Regarding the negation scope detection, they achieved an 84.78 F1 score for a partial match and 83.25 for an exact match.

*5.4.2 Clinical named entity recognition*

One of the most important abilities of bio-NLP is to identify and extract clinical entities within the text. This kind of algorithm (i.e., NER) can support so many types of methods, such as medical concepts extraction, biomedical summarization algorithms, and clinical decision support systems. Souza et al. (2019)[52] describe their preliminary work with promising results on exploring conditional random fields (CRF) algorithms to perform NER in clinical pt-br texts. They used different fragments of our corpus and different annotations granularities (STYs and SGRs) to train and evaluate their model. Considering the best results in the exact-match approach, they achieved a 0.84 F1 score for "Pharmacologic Substance" and 0.71 for "Abbreviation" STYs, which is in line with the IAA scores that we calculated. For the SGRs "Disorder" and "Procedure," they achieved 0.76 and 0.70, respectively.

### 5.5 Additional remarks and future work

Despite the extensive qualitative and quantitative analysis of the results, and recognition of the reproducibility of our corpus, the study limitations and future work need to be discussed. Because they were created from scratch, the guidelines went through a slow process of evolution, which occurred together with the maturation of the annotators, after an extended period of annotation and analysis of new cases. To solve the inconsistencies generated by constant guidelines updates over the course of annotation, Campillos et al. (2018)[4] executed homogenization scripts to track and fix some of these irregularities. We opted to maintain the corpus as it was delivered by the adjudicators, but with the final guidelines and corpus in hands, one can run scripts to harmonize annotations. Following discussions with the annotators, we realized that the annotation task in clinical pt-br texts do not present any additional challenges if we compare with English texts, and this is reinforced by the similar IAA scores with other studies. Furthermore, we concluded that the annotation tool was essential for them with regard to project time constraints. They claimed that annotation suggestions based on previously labeled terms and UMLS API saved them a considerable amount of time. However, analyzing the annotation errors together, we verified that the annotation assistant helped to spread some inconsistencies throughout the corpus. This was because, at some point, the annotation assistant became a very trusted feature for the annotators, occasioning it to be used quickly and carelessly by users, without the annotators checking if the assistant's suggestion was really valid. Additionally, the web-based annotation tool helped us to ease the complex logistics (already discussed by [5]) of training, monitoring and coordinating several annotators at different locations and times.

Thus, some guideline changes that should be followed by all annotators were overwhelmed by suggestions based on annotations made before the update. Therefore, the use of this kind of feature is beneficial, but it should be used carefully. Nevertheless, the UMLS API suggestion feature prevented the annotators from searching for a concept on the UMLS Metathesaurus browser, which was one of the complaints made by Deleger et al. (2012) [37], who indicated that access to an online browser slowed down the annotation process. Another issue common to their study and ours was that, occasionally, the UMLS STY assignment was confusing, as the concept "chronic back pain," is defined as a "Sign or Symptom," but "chronic low back pain" is defined as a "Disease or Syndrome" for example. Additional types for "Negation" and "Abbreviation" annotation gave our corpus an even greater coverage of clinical NLP tasks, as proved by the application of our corpus in a Negation and Scope detection algorithm. Moreover, as a secondary contribution, we were able to build Negation cues and Abbreviations dictionaries for further research, which will be available as the SemClinBr corpus. To replicate this annotation task and address the difficulties associated both with annotation time and guidelines refinement, one could reduce the number of STYs by grouping them into SGRs, as proposed by McCray et al. (2001)[33] and applied in other annotation efforts like Albright et al. (2013)[32].

The recruitment and extensive use of medical students in annotation helped us to finish the annotation considerably faster than if we had depended exclusively on physicians and health professionals due to their availability. In some sense, this reinforces Xia and Yetisgen-Yildiz (2012) claim[14], in which medical training is not the only factor to consider for biomedical annotation tasks. Concerning a complete usefulness assessment of our corpus, we intend to apply it in many other clinical NLP tasks. For sequence labeling tasks, in particular, we want to measure the correlation between algorithm accuracy and IAA for each semantic type. To cover temporal reasoning tasks, we want to expand our annotation schema and create a subset of this corpus with temporal annotation. Furthermore, we need to determine the effect a corpus homogenization process on these NLP/ML algorithms performances.

# 6. Conclusion

We have reported the entire development process of SemClinBr, a semantically annotated corpus for pt-br clinical NLP tasks to provide a common development and evaluation resource for biomedical NLP researchers. To the best of our knowledge, this is the first clinical corpus available for pt-br. Similar annotation projects were surveyed to identify common steps and lessons learned so that mistakes could be avoided and our work could be improved. Furthermore, from this survey, we identified a set of aspects that we cannot find together in other projects (i.e., multi-institutional texts, multiple document types, various medical specialties, high granularity annotation, and a high number of documents), which makes us believe that our annotation task is one of the most complex in clinical NLP literature, considering its generality. We described the data selection, the design of the annotation guidelines (and its refinement process), the annotation tool development, and the annotation workflow. The reliability and usefulness of the corpus were assessed by extensive quantitative and qualitative analysis, including agreement score calculation, error analysis, and application of our corpus in clinical NLP algorithms. Finally, we conclude that the SemClinBr, as well as the developed annotation tool, guidelines, and Negation/Abbreviation dictionaries, could serve as a background for further clinical NLP studies, especially for the Portuguese language.

# 7. Resource availability

The resources produced here (i.e., corpus, annotation guidelines and negation/abbreviation dictionaries) will be made available to the clinical NLP community, with a license agreement regarding scientific and non-commercial use only.

# Acknowledgments

This work was supported by Philips Research North America, Pontifical Catholic University of Paraná, and "*Coordenação de Aperfeiçoamento de Pessoal de Nível Superior - Brasil (CAPES) - Finance Code 001*".

# References

[1]    P. Yadav, M. Steinbach, V. Kumar, G. Simon, Mining Electronic Health Records (EHRs): A Survey, ACM Comput. Surv. 50

(2018) 1–40. doi:10.1145/3127881.

[2] T. Botsis, G. Hartvigsen, F. Chen, C. Weng, Secondary Use of EHR: Data Quality Issues and Informatics Opportunities., Summit on Translat. Bioinforma. 2010 (2010) 1–5. http://www.ncbi.nlm.nih.gov/pubmed/21347133.

[3] A. Névéol, H. Dalianis, S. Velupillai, G. Savova, P. Zweigenbaum, Clinical Natural Language Processing in languages other than English: opportunities and challenges, J. Biomed. Semantics. 9 (2018) 12. doi:10.1186/s13326-018-0179-8.

[4] L. Campillos, L. Deléger, C. Grouin, T. Hamon, A.-L. Ligozat, A. Névéol, A French clinical corpus with comprehensive semantic annotations: development of the Medical Entity and Relation LIMSI annOtated Text corpus (MERLOT), Lang. Resour. Eval. 52 (2018) 571–601. doi:10.1007/s10579-017-9382-y.

[5] A. Roberts, R. Gaizauskas, M. Hepple, G. Demetriou, Y. Guo, I. Roberts, A. Setzer, Building a semantically annotated corpus of clinical texts, J. Biomed. Inform. 42 (2009) 950–966. doi:10.1016/j.jbi.2008.12.013.

[6] Y. Wang, Annotating and recognising named entities in clinical notes, in: Proc. ACL-IJCNLP 2009 Student Res. Work. - ACL-IJCNLP '09, Association for Computational Linguistics, Morristown, NJ, USA, 2009: p. 18. doi:10.3115/1667884.1667888.

[7] Ö. Uzuner, B.R. South, S. Shen, S.L. DuVall, 2010 i2b2/VA challenge on concepts, assertions, and relations in clinical text, J. Am. Med. Informatics Assoc. 18 (2011) 552–556. doi:10.1136/amiajnl-2011-000203.

[8] H. Suominen, S. Salanterä, S. Velupillai, W.W. Chapman, G. Savova, N. Elhadad, S. Pradhan, B.R. South, D.L. Mowery, G.J.F. Jones, J. Leveling, L. Kelly, L. Goeuriot, D. Martinez, G. Zuccon, Overview of the ShARe/CLEF eHealth Evaluation Lab 2013, in: Lect. Notes Comput. Sci. (Including Subser. Lect. Notes Artif. Intell. Lect. Notes Bioinformatics), 2013: pp. 212–231. doi:10.1007/978-3-642-40802-1_24.

[9] R.I. Doğan, R. Leaman, Z. Lu, NCBI disease corpus: A resource for disease name recognition and concept normalization, J. Biomed. Inform. 47 (2014) 1–10. doi:10.1016/j.jbi.2013.12.006.

[10] S. Pradhan, N. Elhadad, W. Chapman, S. Manandhar, G. Savova, SemEval-2014 Task 7: Analysis of Clinical Text, in: Proc. 8th Int. Work. Semant. Eval. (SemEval 2014), Association for Computational Linguistics, Stroudsburg, PA, USA, 2014: pp. 54–62. doi:10.3115/v1/S14-2007.

[11] N. Elhadad, S. Pradhan, S. Gorman, S. Manandhar, W. Chapman, G. Savova, SemEval-2015 Task 14: Analysis of Clinical Text, in: Proc. 9th Int. Work. Semant. Eval. (SemEval 2015), Association for Computational Linguistics, Stroudsburg, PA, USA, 2015: pp. 303–310. doi:10.18653/v1/S15-2051.

[12] A. Stubbs, Ö. Uzuner, Annotating risk factors for heart disease in clinical narratives for diabetic patients, J. Biomed. Inform. 58 (2015) S78–S91. doi:10.1016/j.jbi.2015.05.009.

[13] M. Oronoz, K. Gojenola, A. Pérez, A.D. de Ilarraza, A. Casillas, On the creation of a clinical gold standard corpus in Spanish: Mining adverse drug reactions, J. Biomed. Inform. 56 (2015) 318–332. doi:10.1016/j.jbi.2015.06.016.

[14] F. Xia, M. Yetisgen-Yildiz, Clinical Corpus Annotation: Challenges and Strategies, in: Proc. Third Work. Build. Eval. Resour. Biomed. Text Min. Int. Conf. Lang. Resour. Eval., Istanbul, 2012. http://faculty.washington.edu/melihay/publications/LREC_BioTxtM_2012.pdf.

[15] K. Bretonnel Cohen, D. Demner-Fushman, Biomedical Natural Language Processing, John Benjamins Publishing Company, Amsterdam, 2014. doi:10.1075/nlp.11.

[16] O. Uzuner, Y. Luo, P. Szolovits, Evaluating the State-of-the-Art in Automatic De-identification, J. Am. Med. Informatics Assoc. 14 (2007) 550–563. doi:10.1197/jamia.M2444.

[17] A. Stubbs, C. Kotfila, Ö. Uzuner, Automated systems for the de-identification of longitudinal clinical narratives: Overview of 2014 i2b2/UTHealth shared task Track 1, J. Biomed. Inform. 58 (2015) S11–S19. doi:10.1016/j.jbi.2015.06.007.

[18] O. Uzuner, I. Goldstein, Y. Luo, I. Kohane, Identifying Patient Smoking Status from Medical Discharge Records, J. Am. Med. Informatics Assoc. 15 (2008) 14–24. doi:10.1197/jamia.M2408.

[19] O. Uzuner, Recognizing Obesity and Comorbidities in Sparse Data, J. Am. Med. Informatics Assoc. 16 (2009) 561–570. doi:10.1197/jamia.M3115.

[20] Ö. Uzuner, I. Solti, E. Cadag, Extracting medication information from clinical text, J. Am. Med. Informatics Assoc. 17 (2010)

514–518. doi:10.1136/jamia.2010.003947.

[21] O. Uzuner, A. Bodnari, S. Shen, T. Forbush, J. Pestian, B.R. South, Evaluating the state of the art in coreference resolution for electronic medical records, J. Am. Med. Informatics Assoc. 19 (2012) 786–791. doi:10.1136/amiajnl-2011-000784.

[22] W. Sun, A. Rumshisky, O. Uzuner, Evaluating temporal relations in clinical text: 2012 i2b2 Challenge, J. Am. Med. Informatics Assoc. 20 (2013) 806–813. doi:10.1136/amiajnl-2013-001628.

[23] A. Stubbs, C. Kotfila, H. Xu, Ö. Uzuner, Identifying risk factors for heart disease over time: Overview of 2014 i2b2/UTHealth shared task Track 2, J. Biomed. Inform. 58 (2015) S67–S77. doi:10.1016/j.jbi.2015.07.001.

[24] Ö. Uzuner, I. Solti, F. Xia, E. Cadag, Community annotation experiment for ground truth generation for the i2b2 medication challenge, J. Am. Med. Informatics Assoc. 17 (2010) 519–523. doi:10.1136/jamia.2010.004200.

[25] W. Sun, A. Rumshisky, O. Uzuner, Annotating temporal information in clinical narratives, J. Biomed. Inform. 46 (2013) S5–S12. doi:10.1016/j.jbi.2013.07.004.

[26] A. Stubbs, Ö. Uzuner, Annotating longitudinal clinical narratives for de-identification: The 2014 i2b2/UTHealth corpus, J. Biomed. Inform. 58 (2015) S20–S29. doi:10.1016/j.jbi.2015.07.020.

[27] S. Bethard, G. Savova, W.-T. Chen, L. Derczynski, J. Pustejovsky, M. Verhagen, SemEval-2016 Task 12: Clinical TempEval, in: Proc. 10th Int. Work. Semant. Eval., Association for Computational Linguistics, Stroudsburg, PA, USA, 2016: pp. 1052–1062. doi:10.18653/v1/S16-1165.

[28] S. Bethard, G. Savova, M. Palmer, J. Pustejovsky, SemEval-2017 Task 12: Clinical TempEval, in: Proc. 11th Int. Work. Semant. Eval., Association for Computational Linguistics, Stroudsburg, PA, USA, 2017: pp. 565–572. doi:10.18653/v1/S17-2093.

[29] L. Kelly, L. Goeuriot, H. Suominen, T. Schreck, G. Leroy, D.L. Mowery, S. Velupillai, W.W. Chapman, D. Martinez, G. Zuccon, J. Palotti, Overview of the ShARe/CLEF eHealth Evaluation Lab 2014, in: Lect. Notes Comput. Sci. (Including Subser. Lect. Notes Artif. Intell. Lect. Notes Bioinformatics), 2014: pp. 172–191. doi:10.1007/978-3-319-11382-1_17.

[30] P. Patel, D. Davey, V. Panchal, P. Pathak, Annotation of a Large Clinical Entity Corpus, in: Proc. 2018 Conf. Empir. Methods Nat. Lang. Process., Association for Computational Linguistics, Brussels, Belgium, 2018: pp. 2033–2042. https://www.aclweb.org/anthology/D18-1228.

[31] W.F. Styler, S. Bethard, S. Finan, M. Palmer, S. Pradhan, P.C. de Groen, B. Erickson, T. Miller, C. Lin, G. Savova, J. Pustejovsky, Temporal Annotation in the Clinical Domain, Trans. Assoc. Comput. Linguist. 2 (2014) 143–154. http://www.ncbi.nlm.nih.gov/pubmed/29082229.

[32] D. Albright, A. Lanfranchi, A. Fredriksen, W.F. Styler, C. Warner, J.D. Hwang, J.D. Choi, D. Dligach, R.D. Nielsen, J. Martin, W. Ward, M. Palmer, G.K. Savova, Towards comprehensive syntactic and semantic annotations of the clinical narrative, J. Am. Med. Informatics Assoc. 20 (2013) 922–930. doi:10.1136/amiajnl-2012-001317.

[33] A.T. McCray, A. Burgun, O. Bodenreider, Aggregating UMLS semantic types for reducing conceptual complexity., Stud. Health Technol. Inform. 84 (2001) 216–20. doi:10.3233/978-1-60750-928-8-216.

[34] L. Ferreira, A. Teixeira, J.P. da S. Cunha, Information Extraction from Portuguese Hospital Discharge Letters, in: VI Jornadas En Technol. Del Habl. II Iber. SL Tech Work., 2010: pp. 39–42.

[35] R. Roller, H. Uszkoreit, F. Xu, L. Seiffe, M. Mikhailov, O. Staeck, K. Budde, F. Halleck, D. Schmidt, A fine-grained corpus annotation schema of German nephrology records, in: Proc. Clin. Nat. Lang. Process. Work., The COLING 2016 Organizing Committee, Osaka, Japan, 2016: pp. 69–77. https://www.aclweb.org/anthology/W16-4210.

[36] M. Skeppstedt, M. Kvist, G.H. Nilsson, H. Dalianis, Automatic recognition of disorders, findings, pharmaceuticals and body structures from clinical text: An annotation and machine learning study, J. Biomed. Inform. 49 (2014) 148–158. doi:10.1016/j.jbi.2014.01.012.

[37] L. Deleger, Q. Li, T. Lingren, M. Kaiser, K. Molnar, L. Stoutenborough, M. Kouril, K. Marsolo, I. Solti, Building gold standard corpora for medical natural language processing tasks., AMIA ... Annu. Symp. Proceedings. AMIA Symp. 2012 (2012) 144–53. http://www.pubmedcentral.nih.gov/articlerender.fcgi?artid=PMC3540456.

[38] L. Crible, L. Degand, Reliability vs. granularity in discourse annotation: What is the trade-off?, Corpus Linguist. Linguist.

Theory. 15 (2017) 71–99. doi:10.1515/cllt-2016-0046.

[39] E. Hovy, J. Lavid, Towards a "Science" of Corpus Annotation: A New Methodological Challenge for Corpus Linguistics, Int. J. Transl. 22 (2010) 13–36.

[40] R. Artstein, M. Poesio, Inter-Coder Agreement for Computational Linguistics, Comput. Linguist. 34 (2008) 555–596. doi:10.1162/coli.07-034-R2.

[41] G. Hripcsak, A.S. Rothschild, Agreement, the F-Measure, and Reliability in Information Retrieval, J. Am. Med. Informatics Assoc. 12 (2005) 296–298. doi:10.1197/jamia.M1733.

[42] S. Pradhan, N. Elhadad, B.R. South, D. Martinez, L. Christensen, A. Vogel, H. Suominen, W.W. Chapman, G. Savova, Evaluating the state of the art in disorder recognition and normalization of the clinical narrative, J. Am. Med. Informatics Assoc. 22 (2015) 143–154. doi:10.1136/amiajnl-2013-002544.

[43] J.R. Landis, G.G. Koch, The Measurement of Observer Agreement for Categorical Data, Biometrics. 33 (1977) 159. doi:10.2307/2529310.

[44] D. Reidsma, J. Carletta, Reliability Measurement without Limits, Comput. Linguist. 34 (2008) 319–326. doi:10.1162/coli.2008.34.3.319.

[45] H. Dalianis, Characteristics of Patient Records and Clinical Corpora, in: Clin. Text Min., Springer International Publishing, Cham, 2018: pp. 21–34. doi:10.1007/978-3-319-78503-5_4.

[46] G.H.B. Andrade, L.E.S. e Oliveira, C.M.C. Moro, METODOLOGIAS E FERRAMENTAS PARA ANOTAÇÃO DE NARRATIVAS CLÍNICAS, in: CBIS 2016 - XV Congr. Bras. Informática Em Saúde, Goiânia, 2016: pp. 1031–1040.

[47] L.E.S. Oliveira, C.P. Gebeluca, A.M.P. Silva, C.M.C. Moro, S.A. Hasan, O. Farri, A statistics and UMLS-based tool for assisted semantic annotation of Brazilian clinical documents, in: 2017 IEEE Int. Conf. Bioinforma. Biomed., IEEE, 2017: pp. 1072–1078. doi:10.1109/BIBM.2017.8217805.

[48] S. Boisen, M.R. Crystal, R. Schwartz, R. Stone, R. Weischedel, Annotating Resources for Information Extraction, in: Proc. Second Int. Conf. Lang. Resour. Eval., Athens, 2000: pp. 1211--1214. http://www.lrec-conf.org/proceedings/lrec2000/pdf/263.pdf.

[49] P. Domingos, A few useful things to know about machine learning, Commun. ACM. 55 (2012) 78. doi:10.1145/2347736.2347755.

[50] L. Ferreira, C.T. Oliveira, A. Teixeira, J.P. da S. Cunha, Extracção de Informação de Relatórios Médicos, Linguamática. 1 (2009) 89–102.

[51] C. Dalloux, V. Claveau, N. Grabar, L.E.S. Oliveira, C.M.C. Moro, Y.B. Gumiel, D.R. Carvalho, Cross-lingual and Cross-domain Detection of Negation and of its Scope (in press), J. Nat. Lang. Eng. (2020).

[52] J.V.A. de Souza, Y.B. Gumiel, L.E.S. e Oliveira, C.M.C. Moro, Named Entity Recognition for Clinical Portuguese Corpus with Conditional Random Fields and Semantic Groups, in: An. Do XIX Simpósio Bras. Comput. Apl. à Saúde, Sociedade Brasileira de Computação, Niterói, 2019: pp. 318–323.